# Detecting Depression in Thai Blog Posts: a Dataset and a Baseline

Mika Hämäläinen[1,2], Pattama Patpong[3], Khalid Alnajjar[1,2],
Niko Partanen[2] and Jack Rueter[2]

[1]Rootroo Ltd
[2]University of Helsinki
[3]Mahidol University

[2]`firstname.lastname@helsinki.fi` [3]`ppattama@yahoo.com`

## Abstract

We present the first openly available corpus for detecting depression in Thai. Our corpus is compiled by expert verified cases of depression in several online blogs. We experiment with two different LSTM based models and two different BERT based models. We achieve a 77.53% accuracy with a Thai BERT model in detecting depression. This establishes a good baseline for future researcher on the same corpus. Furthermore, we identify a need for Thai embeddings that have been trained on a more varied corpus than Wikipedia. Our corpus, code and trained models have been released openly on Zenodo.

## 1 Introduction

Depression is a significant public-health problem and one of the leading causes of disease burden worldwide. In 2010, Depression was ranked as the sixth leading causes of disability-adjusted life years in South-East Asia (Murray et al., 2012). Globally, more than 264 million people were affected (James et al., 2018). Depression affects 86 million people in South-East Asia region. At its most severe, depression can lead to suicide, which is the second highest cause of death among 15-29 years old in the region (Sharma, 2017).

Depression is a growing problem in Thailand as the country has a globally high depression rate reaching to 1.7 million people (see Kongsuk et al. 2017). At the same time, people dealing with depression struggle to find help (see Dundon 2006), or worse yet, they might receive insufficient treatment (Lotrakul and Saipanish, 2009). A worst-case scenario, depression can lead to suicide. Depression is one of the leading causes of suicide in Thai adolescents (Sukhawaha and Arunpongpaisal, 2017). In 2019, suicide rate of Thai people aged between 20-29 was 667 persons and aged between 30-39 was 959 persons which was the highest numbers of all aged groups[1].

While Thai is not what we would call low-resourced (see Hämäläinen 2021), there are currently no datasets available for detecting depression in the Thai language. We present a new dataset based on blog posts and verified cases of depressed bloggers. Furthermore, we establish a baseline for further research on the topic in the Thai language. We have published the dataset, code and models presented in this paper openly on Zenodo[2].

## 2 Related work

In this section we present some of the recent related work in more detail. While there have been several digital humanities driven research efforts in better understanding depression in text (Girard and Cohn, 2015; Abd Yusof et al., 2017; Loveys et al., 2018), we only focus on work that has been conducted on depression detection. There are also several approaches to depression detection in other languages (Pirina and Çöltekin, 2018; Husseini Orabi et al., 2018; Song et al., 2018).

An onset of major depression can be characterized and detected by investigating social media data such as Twitter (De Choudhury et al., 2013b,a). By exploring a large corpus of Twitter posting by using crowd-sourcing methodology. An SVM classifier was built by divesting a variety of social media measures such as social activity, egonetwork, style, user engagement, emotion, and language. The classifier model predicted with high accuracy (70% and 73 % respectively) predicting ahead of the reported onset of depression and whether or not a post on Twitter could be depressive-indicative postings. Depression levels is detected by a proposed social media depression index.

---

[1]Department of Mental Health report https://dmh.go.th/report/suicide/age.asp
[2]https://zenodo.org/record/4734552

Computerized analysis of various kind of texts related to depression reveals signals of psychiatric disorders. Depression is measured by self-reported symptoms (Rude et al., 2002), by clinical interview (Rude et al., 2003). The Scrambled Sentences Test (SST) (Wenzlaff, 1993) was used to measure of cognitive processing bias of a large sample of college students. Negative cognitive processing biases in resolving ambiguous verbal information can predict depression and subsequent depression symptoms.

Depression has been detected automatically before in Thai Facebook users (Katchapakirin et al., 2018). The authors train several models on RapidMiner. The models rely on metadata for activity on Facebook such as the number of posts posted on a given week day, the number of day-time and night-time posts and the number of shared posts. The authors did not train the models on text, but rather used numerical features extracted from Facebook posts such as the number of first person pronouns and number of positive words.

A screening text-based classification model was also applied to Thai Facebook posts to detect depressive disorder (Hemtanon and Kittiphattanabawon, 2019). Similarly to the previous approach, the authors apply several traditional machine learning techniques to Thai social media text with pre-extracted features.

Kumnunt and Sornil, 2020 present a CNN (convolutional neural network) based approach to depression detection in Thai social media posts. The authors crawl posts tagged with a depression hashtag to build their dataset.

Contrary to any of the current work on Thai depression detection, our approach does not deal with social media posts but lengthier blog posts. We also make our dataset available unlike the existing work.

## 3 Data

Our data was taken from blogs or weblogs. As defined by Mishne (2005), blogs are personal, highly opinionated logs, available publicly, with a subjective writing style, and the use of non-content features such as emotions. Blogs can be used as data sources to predict the most likely state of mind with which the post was written: whether the blogger was depressed, cheerful or bored. Blogs prompt many bloggers to indicate their mood at the time of posting (Mishne et al., 2006).

We have obtained post data from two main platforms: the weblog entitled "Storylog" and personal blogs posted via two blog platforms "bloggang.com" and "blogspot.com". Storylog is a social writing platform in Thai language. It was launched in 2014 (https://storylog.co). Its philosophy is "Everyone has something to tell." Stories shared on it are text-based articles about life experiences, opinions, ideas, as well as inspirations. Posts are organized by the bloggers' areas of interest experience, emotion, inspiration, and short story. Towards the other end of communication, blog readers can comment on the given post.

We use these platforms to collect a corpus of posts that are written by depressed individuals. Only posts, not user comments, are analyzed in this study. Criteria for data collection included:

- each blogger has posted at least 15 post entries
- posts contain depressive experiences indicated by the key words "depressed", "depression", "depressive disorder", "useless", "fail", "death", "overdose", "suicide", "cut", "self harm"
- the content of the post contains one of the three types of depression symptoms (Cassano and Fava, 2002): psychological symptoms, behavioral symptoms, and physical symptoms
- only posts written in prose format are investigated
- poetic format and English translation are excluded

Through manual analysis and verification, we reach to 400 depressed posts published on Storylog between 2015-2019, and 72 personal blogs posts published during the same time period. In total, our corpus contains 472 verified depressed blog posts. To make the corpus viable for machine learning, we further crawl an additional 472 blog posts that do not exhibit depression from the Storylog platform.

| Sentence | Depression |
|---|---|
| การวิ่งหนีความรู้สึกที่ "ยาก" เกินไปทำให้เราปลอดภัยและไม่เจ็บปวด Running away from feeling "difficult" keeps us safe and away from pain. | True |
| ม่ต้องอ่านถ้าไม่อยากฟังรสขมสุดตีนของชีวิตคนคนนึง You don't have to read if you don't want to hear the bitter taste of a person's life. | True |
| แต่ยังจำความเจ็บจากการคาดหวังมากเกินไปได้ดี But I still remember the pain of expecting too much | True |
| เป็นวันที่เหนื่อยจริงๆเหนื่อยทั้งกายและใจมากๆ It was a really tiring day, I am very tired both physically and mentally. | True |
| อืม นั่นล่ะ ความรู้สึกที่ชัดเจนที่สุดมาตั้งนานแล้ว.. Well, that's it. The clearest feeling for a long time. | True |
| เริ่มต้นขึ้นเมื่อเขาใช้เวลาว่างจากการอ่านหนังสืออย่างหนักหน่วงในช่วงวัยรุ่น It began when he spent his free time reading heavily during his teenage years. | False |
| ไม่คิดจะหยิบหนังสือขึ้นมาอ่านหน่อยหรือ Don't you want to pick up and read a book? | False |
| ถ้ามีโอกาสหนูอยากให้พี่เข้ามาอ่านนะหนูอยากไปต่อกับพี่ If there's a chance I want you to come in and read it, I want to go with you. | False |
| ฉันเจอเธอทุกวันที่ป้ายรถประจำทางตอนเช้า I see her every day at the bus stop in the morning. | False |
| เขามีเพียงรอยยิ้มประดับเล็กน้อยบนแก้ม He only has a slightly embellished smile on his cheeks. | False |

Table 1: Example sentences from the dataset and their translations in English

|  | train | valid | test |
|---|---|---|---|
| depressed | 12837 | 1712 | 2567 |
| non-depressed | 12240 | 1632 | 2448 |

Table 2: Number of sentences in the data splits

In total, our dataset has 472 depressed and 472 non-depressed blog posts. We shuffle the data on a sentence level and split it to 75% training, 10 % validation and 15 % testing. The splits are shared across all the models. Table 1 shows some examples of the dataset with translations into English. The number of sentences for training, testing and validation can be seen in Table 2

Words are typically written together without spaces in Thai language and white-spaces are used to indicate sentence boundaries. This means that the blog posts need to be tokenized into sentences and words. We do these tokenizations by using PyThaiNLP (Phatthiyaphaibun et al., 2016).

## 4 Detecting depression

In this section, we describe the four different models we use to automatically detect depression in our dataset. Following Hämäläinen et al. (2021), two of the models are based on LSTM (long short-term memory) models and two on transfer learning on pretrained BERT models (Devlin et al., 2019).

We train our first model using a bi-directional LSTM based model (Hochreiter and Schmidhuber, 1997) using OpenNMT (Klein et al., 2017) with the default settings except for the encoder where we use a BRNN (bi-directional recurrent neural network) (Schuster and Paliwal, 1997) instead of the default RNN (recurrent neural network). We use the default of two layers for both the encoder and the decoder and the default attention model, which is the general global attention presented by Luong et al. (2015). The model is trained for the default 100,000 steps. The model is trained with tokenized sentences as its source and the depression label as its target.

We train an additional LSTM model with the same configuration and the same random seed (3435) with the only difference being that we use pretrained word2vec embeddings for the encoder. We use the Thai2vec embeddings provided as a part of PyThaiNLP[3]. The vector size is 300 and the model has been trained on Thai Wikipedia.

In addition, we train two different BERT based sequence classification models based on the Thai BERT model[4] and Multilingual BERT (Devlin et al., 2019), which has been trained on multiple languages, Thai being one of the latest additions to the cased model. We use the transformers package (Wolf et al., 2020) to conduct the fine tuning with our dataset. The models are fine tuned to predict the depression label from a tokenized sentence. For the Multilingual BERT model, we use the same tokenization as for the other models, but for the Thai BERT model, we run the model's own tokenizier on an untokenized input. As hyperparameters for the fine tuning, we use 3 epochs with 500 warm-up steps for the learning rate scheduler and 0.01 as the strength of the weight decay.

This setup takes into account the current state of the art at the field, and uses recently created resources such as Thai BERT model, with our own custom made dataset. Everything is set up in an easily replicable manner, which ensures that our experiments and results can be used in further work on this important topic.

## 5 Results

In this section, we show and discuss the results obtained with our 4 different models: 2 LSTM based models and 2 BERT based ones. The results of the models can be seen in Table 3.

|  | overall acc | depression acc | no depression acc |
|---|---|---|---|
| LSTM | 76.19% | 77.05% | 75.29% |
| LSTM +w2v | 75.69% | 74.44% | **77.00%** |
| Multilingual BERT | 75.06% | 79.11% | 70.80% |
| Thai BERT | **77.53%** | **79.70%** | 75.25% |

Table 3: The accuracies obtained by the different models

The model finetuned on Thai BERT reaches the highest overall accuracy and the highest accuracy in detecting depressed sentences correctly. However, it misclassifies non depressed sentences as depressed more frequently than the LSTM based model that is trained with word2vec embeddings. This is interesting as the BERT model is superior to the LSTM model in the overall and depression accuracies and notably inferior in the non-depression accuracy.

What makes this result even more interesting is that both the Thai BERT and the pretraiend word2vec model used with the LSTM model have been trained on Thai Wikipedia. This is very likely due to the fact that Wikipedia represents the type of genre that does not exhibit any features of depression, and therefore a word2vec based model that cannot capture contextual information is more likely to label sentences as non depressed leading to a lower accuracy in the depressed label and a higher accuracy in the non depressed label. The contextuality of the BERT embeddings makes it possible for them to be fine tuned better towards the new genre of depressed text.

## 6 Discussion and conclusions

In this paper, we have presented a new expert curated dataset for depression detection in Thai blog posts. Our sentence level results are promising and we are sure we can use our models to snowball more depressed blog posts from platforms such as Storylog for further linguistic analysis as a part of our interdisciplinary research project.

Based on our results, we can identify one easy future direction for enhancing the results obtained by our models. Currently, all freely available pretrained Thai embeddings have been trained on Wikipedia. This is not optimal for several reasons, one being the encyclopedic genre of Wikipedia, the other being the fact that while Wikipedia is written in formal "correct" Thai, blog posts are written in a more colloquial language variety. This means that the vocabulary coverage of Wikipedia data is poor when compared to blog posts.

Our blog corpus consists of 21,002 unique tokens while the word2vec model trained on Wikipedia has embeddings for 51,358 words.

---
[3] https://pythainlp.github.io/tutorials/notebooks/word2vec_examples.html
[4] https://github.com/ThAIKeras/bert

The blog corpus (training, testing and validation combined) contains 6,488 words that are not present in the word2vec model, this means that around 31% of the words present in our blog depression corpus are simply not in a Wikipedia based model. In the future, it is clear that Thai language calls for openly available models that are trained on a larger and more varied internet corpus than solely on Wikipedia.

# References

Noor Fazilla Abd Yusof, Chenghua Lin, and Frank Guerin. 2017. Analysing the causes of depressed mood from depression vulnerable individuals. In *Proceedings of the International Workshop on Digital Disease Detection using Social Media 2017 (DDDSM-2017)*, pages 9–17, Taipei, Taiwan. Association for Computational Linguistics.

Paolo Cassano and Maurizio Fava. 2002. Depression and public health: an overview. *Journal of psychosomatic research*, 53(4):849–857.

Munmun De Choudhury, Scott Counts, and Eric Horvitz. 2013a. Social media as a measurement tool of depression in populations. In *Proceedings of the 5th annual ACM web science conference*, pages 47–56.

Munmun De Choudhury, Michael Gamon, Scott Counts, and Eric Horvitz. 2013b. Predicting depression via social media. In *Proceedings of the International AAAI Conference on Web and Social Media*, volume 7.

Jacob Devlin, Ming-Wei Chang, Kenton Lee, and Kristina Toutanova. 2019. BERT: Pre-training of deep bidirectional transformers for language understanding. In *Proceedings of the 2019 Conference of the North American Chapter of the Association for Computational Linguistics: Human Language Technologies, Volume 1 (Long and Short Papers)*, pages 4171–4186, Minneapolis, Minnesota. Association for Computational Linguistics.

Edith "Emma" Dundon. 2006. Adolescent depression: A metasynthesis. *Journal of Pediatric Health Care*, 20(6):384–392.

Jeffrey M Girard and Jeffrey F Cohn. 2015. Automated audiovisual depression analysis. *Current opinion in psychology*, 4:75–79.

Mika Hämäläinen. 2021. Endangered languages are not low-resourced! In *Multilingual Facilitation*, pages 1–11. University of Helsinki.

Mika Hämäläinen, Khalid Alnajjar, Niko Partanen, and Jack Rueter. 2021. Never guess what I heard... rumor detection in Finnish news: a dataset and a baseline. In *Proceedings of the Fourth Workshop on NLP for Internet Freedom: Censorship, Disinformation, and Propaganda*, pages 39–44, Online. Association for Computational Linguistics.

Siranuch Hemtanon and Nichnan Kittiphattanabawon. 2019. An automatic screening for major depressive disorder from social media in thailand. *รายงานการประชุม วิชาการ เสนอ ผล งาน วิจัย ระดับ ชาติ และ นานาชาติ*, 1(10):103–113.

Sepp Hochreiter and Jürgen Schmidhuber. 1997. Long short-term memory. *Neural computation*, 9(8):1735–1780.

Ahmed Husseini Orabi, Prasadith Buddhitha, Mahmoud Husseini Orabi, and Diana Inkpen. 2018. Deep learning for depression detection of Twitter users. In *Proceedings of the Fifth Workshop on Computational Linguistics and Clinical Psychology: From Keyboard to Clinic*, pages 88–97, New Orleans, LA. Association for Computational Linguistics.

Spencer L James, Degu Abate, Kalkidan Hassen Abate, Solomon M Abay, Cristiana Abbafati, Nooshin Abbasi, Hedayat Abbastabar, Foad Abd-Allah, Jemal Abdela, Ahmed Abdelalim, et al. 2018. Global, regional, and national incidence, prevalence, and years lived with disability for 354 diseases and injuries for 195 countries and territories, 1990–2017: a systematic analysis for the global burden of disease study 2017. *The Lancet*, 392(10159):1789–1858.

Kantinee Katchapakirin, Konlakorn Wongpatikaseree, Panida Yomaboot, and Yongyos Kaewpitakkun. 2018. Facebook social media for depression detection in the thai community. In *2018 15th International Joint Conference on Computer Science and Software Engineering (JCSSE)*, pages 1–6. IEEE.

Guillaume Klein, Yoon Kim, Yuntian Deng, Jean Senellart, and Alexander M. Rush. 2017. OpenNMT: Open-Source Toolkit for Neural Machine Translation. In *Proc. ACL*.

Thoranin Kongsuk, Suttha Supanya, Kedsaraporn Kenbubpha, Supranee Phimtra, Supattra Sukhawaha, Jintana Leejongpermpoon, et al. 2017. Services for depression and suicide in thailand. *WHO South-East Asia journal of public health*, 6(1):34.

Boriharn Kumnunt and Ohm Sornil. 2020. Detection of depression in thai social media messages using deep learning. In *Proceedings of the 1st International Conference on Deep Learning Theory and Applications – DeLTA*, pages 111–118.

Manote Lotrakul and Ratana Saipanish. 2009. How do general practitioners in thailand diagnose and treat patients presenting with anxiety and depression? *Psychiatry and clinical neurosciences*, 63(1):37–42.

Kate Loveys, Jonathan Torrez, Alex Fine, Glen Moriarty, and Glen Coppersmith. 2018. Cross-cultural differences in language markers of depression online. In *Proceedings of the Fifth Workshop on Computational Linguistics and Clinical Psychology: From Keyboard to Clinic*, pages 78–87, New Orleans, LA. Association for Computational Linguistics.

Minh-Thang Luong, Hieu Pham, and Christopher D Manning. 2015. Effective approaches to attention-based neural machine translation. *arXiv preprint arXiv:1508.04025*.

Gilad Mishne. 2005. Experiments with mood classification in blog posts. In *Proceedings of ACM SIGIR 2005 workshop on stylistic analysis of text for information access*, volume 19, pages 321–327. Citeseer.

Gilad Mishne, Maarten De Rijke, et al. 2006. Capturing global mood levels using blog posts. In *AAAI spring symposium: computational approaches to analyzing weblogs*, volume 6, pages 145–152.

Christopher JL Murray, Theo Vos, Rafael Lozano, Mohsen Naghavi, Abraham D Flaxman, Catherine Michaud, Majid Ezzati, Kenji Shibuya, Joshua A Salomon, Safa Abdalla, et al. 2012. Disability-adjusted life years (dalys) for 291 diseases and injuries in 21 regions, 1990–2010: a systematic analysis for the global burden of disease study 2010. *The lancet*, 380(9859):2197–2223.

Wannaphong Phatthiyaphaibun, Korakot Chaovavanich, Charin Polpanumas, Arthit Suriyawongkul, Lalita Lowphansirikul, and Pattarawat Chormai. 2016. PyThaiNLP: Thai Natural Language Processing in Python. *Zenodo*.

Inna Pirina and Çağrı Çöltekin. 2018. Identifying depression on Reddit: The effect of training data. In *Proceedings of the 2018 EMNLP Workshop SMM4H: The 3rd Social Media Mining for Health Applications Workshop & Shared Task*, pages 9–12, Brussels, Belgium. Association for Computational Linguistics.

Stephanie S Rude, Carmen R Valdez, Susan Odom, and Arshia Ebrahimi. 2003. Negative cognitive biases predict subsequent depression. *Cognitive Therapy and Research*, 27(4):415–429.

Stephanie S Rude, Richard M Wenzlaff, Bryce Gibbs, Jennifer Vane, and Tavia Whitney. 2002. Negative processing biases predict subsequent depressive symptoms. *Cognition & Emotion*, 16(3):423–440.

Mike Schuster and Kuldip K Paliwal. 1997. Bidirectional recurrent neural networks. *IEEE transactions on Signal Processing*, 45(11):2673–2681.

Shamila Sharma. 2017. Talk about depression, strengthen depression-related services who. In *World Health Organization*.

Hoyun Song, Jinseon You, Jin-Woo Chung, and Jong C. Park. 2018. Feature attention network: Interpretable depression detection from social media. In *Proceedings of the 32nd Pacific Asia Conference on Language, Information and Computation*, Hong Kong. Association for Computational Linguistics.

S Sukhawaha and S Arunpongpaisal. 2017. Risk factor and suicide theory associated with suicide in adolescents: A narrative reviews. *Journal of the Psychiatric Association of Thailand*, 62(4):359–378.

Richard M Wenzlaff. 1993. The mental control of depression: Psychological obstacles to emotional well-being. *Century psychology series. Handbook of mental control*.

Thomas Wolf, Lysandre Debut, Victor Sanh, Julien Chaumond, Clement Delangue, Anthony Moi, Pierric Cistac, Tim Rault, Rémi Louf, Morgan Funtowicz, Joe Davison, Sam Shleifer, Patrick von Platen, Clara Ma, Yacine Jernite, Julien Plu, Canwen Xu, Teven Le Scao, Sylvain Gugger, Mariama Drame, Quentin Lhoest, and Alexander M. Rush. 2020. Transformers: State-of-the-art natural language processing. In *Proceedings of the 2020 Conference on Empirical Methods in Natural Language Processing: System Demonstrations*, pages 38–45, Online. Association for Computational Linguistics.